\documentclass[preprint,1p,times]{elsarticle}%
\usepackage[utf8]{inputenc}
\usepackage[english]{babel} 
\usepackage{microtype}
\usepackage{graphicx}
\usepackage{subfigure}
\usepackage{booktabs} 
\usepackage{tipa}
\usepackage{upgreek}
\usepackage{lipsum}

\usepackage{mathalpha}

\usepackage{amsmath,amssymb}
\usepackage{dsfont}
\usepackage{enumerate}
\usepackage{graphicx}      
\usepackage{comment}
\usepackage{algorithmic}
\usepackage{amsthm}

\DeclareMathAlphabet{\mathpzc}{OT1}{pzc}{m}{it}

\usepackage{mathtools}
\usepackage[ruled,vlined]{algorithm2e}
\usepackage{xcolor}

\begin{document}
\begin{frontmatter}

\title{Constrained Model-Free Reinforcement Learning for Process Optimization}

\author[CPSE]{Elton Pan}
\author[UCL]{Panagiotis Petsagkourakis\corref{cor1}}
 \ead{p.petsagkourakis@ucl.ac.uk}
 \cortext[cor1]{Corresponding authors}
\author[MU]{Max Mowbray}
\author[MU]{Dongda Zhang}
\author[CPSE]{Ehecatl Antonio del Rio-Chanona\corref{cor1}}
 \ead{a.del-rio-chanona@imperial.ac.uk}

\address[CPSE]{Centre for Process Systems Engineering, Department of Chemical Engineering, Imperial College London, UK}
\address[UCL]{Centre for Process Systems Engineering, Department of Chemical Engineering, University College London, UK}
\address[MU]{Department of Chemical Engineering and Analytical Science, University of Manchester, UK}

\date{November 2020}

\begin{abstract}                

  Reinforcement learning (RL) is a control approach that can handle nonlinear stochastic optimal control problems. However, despite the promise exhibited, RL has yet to see marked translation to industrial practice primarily due to its inability to satisfy state constraints. In this work we aim to address this challenge. We propose an “oracle”-assisted constrained Q-learning algorithm that guarantees the satisfaction of joint chance constraints with a high probability, which is crucial for safety critical tasks. To achieve this, constraint tightening (backoffs) are introduced and adjusted using Broyden’s method, hence making the backoffs self-tuned. This results in a methodology that can be imbued into RL algorithms to ensure constraint satisfaction. We analyze the performance of the proposed approach and compare against nonlinear model predictive control (NMPC). The favorable performance of this algorithm signifies a step towards the incorporation of RL into real world optimization and control of engineering systems, where constraints are essential.
\end{abstract}
\end{frontmatter}


\section{Introduction}
The optimization of nonlinear stochastic processes poses a challenge for conventional control schemes given the requirement of an accurate process model and a method to simultaneously handle process stochasticity and satisfy state and safety constraints. Recent works have explored the application of model-free Reinforcement Learning (RL) methods for dynamic optimization of batch processes within the chemical and biochemical industries \cite{petsagkourakis2020reinforcement,singh2020reinforcement}. Many of these demonstrate the capability of RL algorithms to learn a control law independently from a nominal process model, but negate proper satisfaction of state and safety constraints. In this work, we propose constrained Q-learning, a model-free algorithm to meet the operational and safety requirements of constraint satisfaction with high probability. 

\subsection{Model-Free Reinforcement Learning}

RL encompasses a subfield of machine learning, which aims to learn an optimal policy for a system that can be described as a Markov decision process (MDP). Importantly, MDPs assume the Markov property, such that the future transitions (dynamics) of the process are only dependent upon the current state and control action, and not upon the process history \cite{szepesvari2010algorithms}. 

Dynamic programming (DP) methods provide exact solution to MDPs under knowledge of the exact process dynamics \cite{bertsekas1995dynamic} . A subset of RL algorithms are known as action-value (or Q-learning) methods. These methods are closely related to DP, but instead, learn an approximate parameterization of the optimal action-value function independently of explicit knowledge of the exact dynamics \cite{sutton2018reinforcement}. This is achieved by sampling the response of the underlying Markov process, eliminating the requirement for explicit assumption regarding the stochastic nature of the system. This is a particularly powerful concept in the domain of process control and optimization, given the inherent uncertainties and (slow) non-stationary dynamics often characteristic of process systems \cite{spielberg2019toward, petsagkourakis2020reinforcement}. 

\begin{figure}[h!]
\centering
    \includegraphics[scale=0.55]{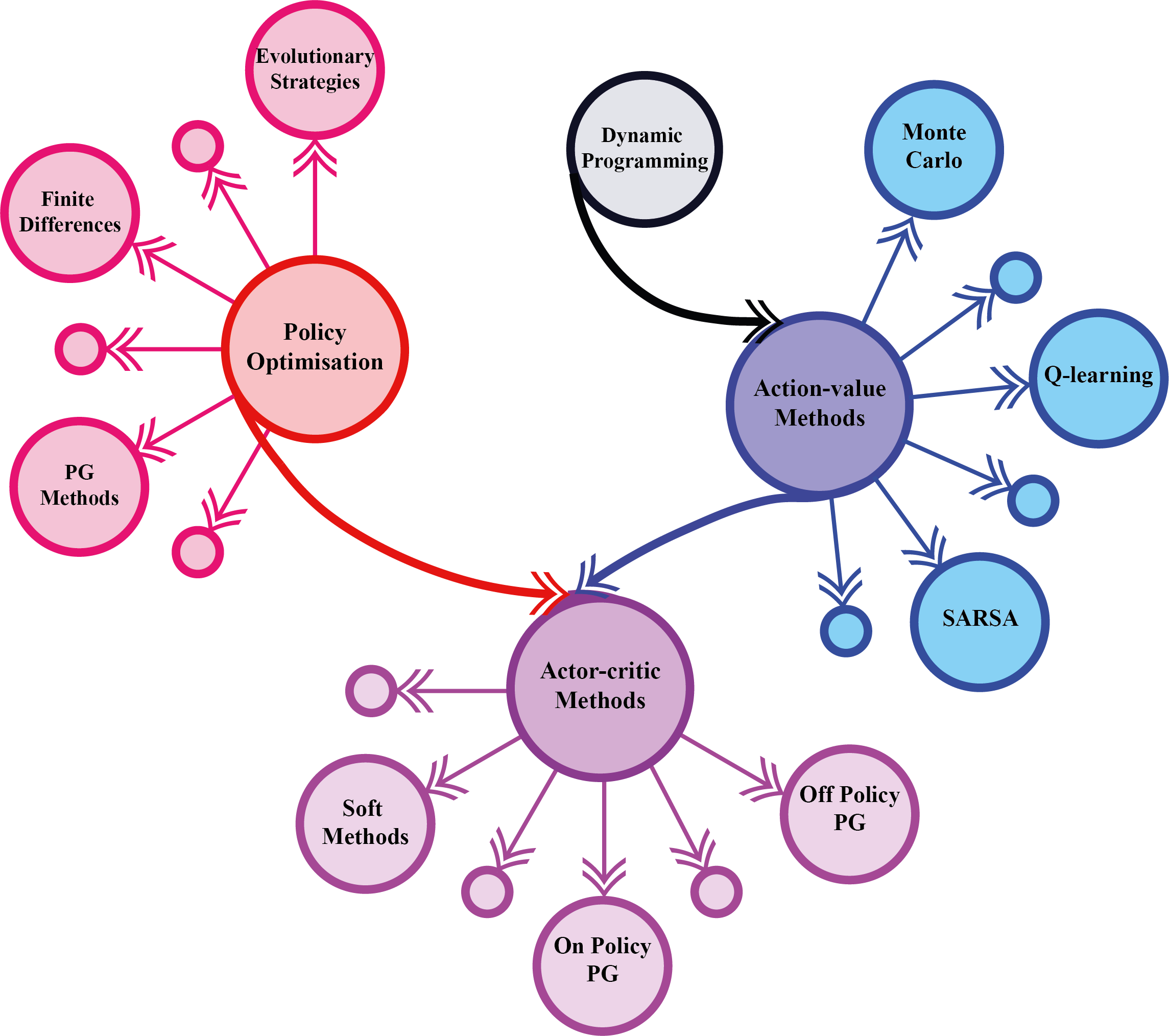}
    \caption{ {The landscape of model free reinforcement learning (RL). Model free RL can be broadly constituted by policy optimization and action-value methods with the intersection of the two characterised by actor-critic methods. The figure does not exhaustively detail the different algorithms, but rather broadly describes those approaches that are dominant within each respective class. The use of bubbles with no description denotes those algorithms not detailed for the sake of conciseness.}}
    \label{fig:RLlandscape}
\end{figure}

The field of model-free reinforcement learning extends beyond action-value methods; two other approaches exist in the form of policy optimization \cite{lehman2018more, sutton2000policy} and actor-critic \cite{haarnoja2018soft, lillicrap2015continuous} methods. { The relationships between the different sub-classes of RL algorithms are expressed by Fig. \ref{fig:RLlandscape}. The following analysis proceeds with reference to that Figure. Action}-value methods explicitly learn a parameterization of the action-value function, whereas policy optimization methods implicitly learn the value space and instead learn and parameterize a policy directly \cite{shin2019reinforcement}. Typically, policy optimization approaches deploy a Monte Carlo method to gain estimate of the value-function corresponding to the current policy parameters. This provides a search direction for further policy improvement. However, policy optimization methods tend to follow \textit{on-policy} learning rules, which means that data collected under a given policy may only be used for one learning update before being discarded \cite{sutton2018reinforcement}. On-policy learning, combined with dependency on the use of a Monte Carlo method for evaluation of the search direction, provides significant sample complexity. Actor-critic methods partially reduce the associated sample complexity by explicitly learning a parameterization of the action-value function (the critic) as well as a policy (the actor) - removing dependency on Monte Carlo sampling \cite{szepesvari2010algorithms}. In essence, the parameterization of the action-value function enables conceptualisation of actor-critic methods as those algorithms at the intersection of policy optimisation and action-value methods. Via this analysis, the apparent benefit of action-value methods is of reduced sample-complexity. This is further improved when experience replay is integrated \cite{rolnick2019experience}, enabling saving of data observed in previous simulations for future learning updates. As such, this directs focus in the analysis provided by the following section, where review of action-value function approximation in chemical engineering is presented.


\subsection{Action-Value Function Approximation in Chemical Engineering}
In recent years, there has been a growing interest in the development of RL controllers in the domain of chemical and biochemical processing and this is reflected by the rapid growth of literature in the area. {The interest is primarily generated by two attractive properties - reduced sample complexity and the ability to integrate experience replay \cite{RLstreamorsave}. There are two classes of RL algorithms which leverage value function approximation: actor-critic, and action-value methods. The use of the two approaches is reviewed in the following discussion.}

Actor-critic methods have been applied widely, given their explicit parameterisation of a policy (actor) and the associated sample efficiency induced via parameterisation of a value function (critic). In \cite{goulart2020nautonomous} an actor-critic method for pH control of wastewater from an industrial electroplating process is proposed and benchmarked against a proportional-integral-derivative (PID) controller. Similarly, in \cite{lawrence2020optimal}, the authors present an actor-critic based RL framework for optimal PID controller tuning, including a mechanism for \textit{antiwindup}. In \cite{xie2020model}, the authors propose a framework to augment the actor-critic algorithm (Deep deterministic policy gradient, DDPG) with nonlinear model predictive control (NMPC). The framework is then demonstrated on two case studies, both of which highlight the framework's ability to provide marked improvements in the efficacy of offline policy training, relative to a vanilla implementation of DDPG. In \cite{spielberg2019toward}, the authors identify the potential benefits of a data-driven RL controller with respect to ease of system re-identification and demonstrate the application of an actor-critic algorithm to a number of case studies including control of a high purity distillation column. 

For action-value methods to be deployed in continuous control spaces, they typically require augmentation with a further optimization method for determining the optimal control \cite{ryu2019caql}, and currently, very efficient optimization algorithms exist \cite{wachter2002interior,nocedal2006numerical} to facilitate this. Action-value methods are particularly attractive given they generally encompass the most sample efficient of algorithms in the RL toolbox. In \cite{sajedian2019optimisation}, a Q-learning approach was applied for nanostructure and nanosurface design. Similarly, \cite{zhou2019optimization} proposes a Q-learning method for the purpose of molecular design. In \cite{hwangbo2020design}, the authors integrate a Q-learning method with Aspen Plus for control of a downstream separation process, demonstrating improved performance relative to open loop operation. In \cite{lee2006approximate} an action-value method is proposed in the context of process scheduling and in \cite{lee2005approximate} the authors demonstrate the application of action-value methods for the control of stochastic, nonlinear processes. The above have shown good success in chemical process optimization and control {providing basis for further use of action-value methods and development of 'saving focused' learning strategies.} However, for RL methods to be deployed in many instances of process engineering, they must satisfy constraints (with high probability in the stochastic setting). This is the challenge that the current work addresses.

\subsection{Related Work}

As highlighted in \cite{leurent2020robustadaptive}, there exists relative inertia in application of RL to industrial control problems. Specifically, in the chemical and biochemical process industries, the development of RL methods to guarantee safe process operation and constraint satisfaction would enhance prospective deployment of RL-based systems in the scope of optimization and control. The literature documents a number of approaches to safe constraint satisfaction that typically augment a pure model-free RL-based controller, with a separate system that has basis in direct optimal control. Such augmented systems are broadly constituted by barrier function \cite{choi2020reinforcement,taylor2020learning,cheng2019end} and safety filter methods \cite{Safefilter,wabersich2018safe}. The deployment of these approaches dictates a nominal description of the process dynamics, method to handle process stochasticity and often impose nontrivial policy or value learning rules. The aspect of model dependency particularly dampens the initial attraction of an RL approach within the context of process control. 

Other works have explored the development of methods for safe constraint satisfaction by leveraging the value framework provided by MDPs, preserving performance independent of a process model. These methods either add penalty to the original reward function (objective) for constraint violation \cite{lee2005approximate,tessler2018reward} or augment the original MDP to take the form of a constrained MDP (CMDP) \cite{altman1999constrained}. If approached crudely, the former approach introduces a number of hyperparmeters, which are typically chosen on the basis of heuristics and have bearing on policy optimality. This is also discussed in \cite{achiam2017constrained, engstrom2020implementation}. The latter approach is underpinned by the learning of surrogate cost functions for each individual constraint combined with appropriate adaptation of the policy \cite{achiam2017constrained} or value learning rule. Both approaches ensure constraint satisfaction only in expectation \cite{satija2020constrained}, which is insufficient for control and optimization of (bio)chemical processes. As most engineering systems are safety critical, satisfaction of constraints with high probability is a necessity. {In view of this problem, \cite{russel2020robust} present an approach which combines the CMDP concept with consideration of worst case realisations of process uncertainty, in an attempt to robustify the framework. However, the approach is only demonstrated empirically in a discrete domain (almost all industrial problems are continuous)
}.  In the same rationale, a Lyapunov-based approach is proposed in \cite{Chow2019}, where a Lyapunov function is found and the unconstrained policy is projected to a safety layer allowing the satisfaction of constraints in expectation. However this is not the optimal trade-off, as the closest action is not necessarily optimal. Additionally, the satisfaction of constraints in expectation is not sufficient for the most real-world applications. A similar approach is also presented in \cite{huh2020safe}, where a lyapunov function is identified, but for the purpose of safe exploration. This is not necessarily directly applicable to the process industries, where online exploratory control decisions pose high operational risk and the fact that operational constraints are not explicitly considered.
 {More recently, work has been proposed in \cite{peng2021separated}, based on the augmented lagrangian. However, the method negates the inclusion of state and control dependency within the penalty term, instead appending the penalty in a crude form of credit assignment, therefore implicitly neglecting an aspect of the RL problem. It is likely that this approach leads to conservative control policies. The method presented in \cite{petsagkourakis2020chance}, provides a mechanism for the state and control dependency desired. The work presented here, leverages similar concepts but in the context of action-value function learning, improving the data-efficiency of the algorithm alongside other advantages discussed herein.}

\subsection{Contribution}
To our knowledge, no approach has been proposed which achieves constraint satisfaction with high probability for action-value methods. In this work, we propose an action-value method, which guarantees constraint satisfaction with high probability. Here, we first learn an unconstrained actor and surrogate constraint action-value functions. We then subsequently construct a constrained action-value function as a superimposition of the unconstrained actor with the surrogate constraints (in the form of a constrained optimization problem). The constrained actor is iteratively tuned, as learning proceeds, via localised backoffs \cite{bradford2020stochastic, mehta2016integration, rafiei2020trust} to penalize constraint violation. Conceptually, backoffs provide a policy variant shaping mechanism to ensure high probability satisfaction \cite{ng1999policy}. Tuning comprises a Monte Carlo method to estimate the probability of constraint violation under the policy combined with Broyden's root finding method. Specifically, Broyden's method is used to update the backoff values. {Importantly, the dimensionality of the tuning problem is equivalent to the number of operational constraints imposed upon the problem providing a scale-able algorithm.} Given the constrained actor action-value function and the fast inference associated with neural networks, efficient optimization strategies may be deployed for determination of the optimal control. {Further, the algorithm proposed incorporates the use of an experience replay memory store, promoting interpretation of the method as a saving algorithm \cite{RLstreamorsave}. This is particularly important in the scope of continual learning and improvement of the policy once deployed to the real process \cite{rolnick2019experience}.}  The work is arranged as follows; the problem description is formalized in section 2, the methodology proposed in section 3 and demonstrated empirically in section 4 via two benchmark case studies. 


\section{Problem Statement}

\subsection{Reinforcement Learning in Process Engineering}
\begin{figure}[h!]
    \centering
    \includegraphics[scale=0.37]{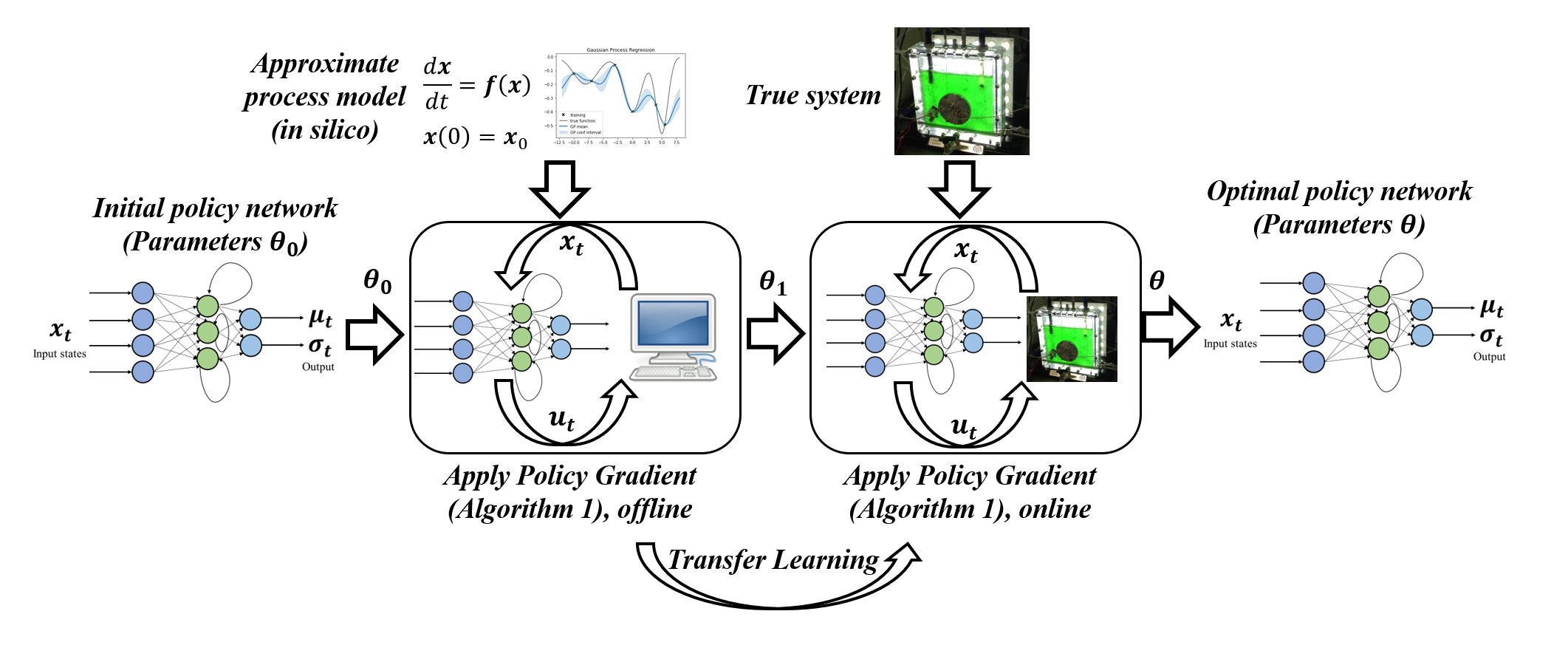}
    \caption{Schematic representation of RL for chemical process optimization (Adapted from \cite{petsagkourakis2020reinforcement}).}
    \label{RL_scheme}
\end{figure}

 {Using reinforcement learning directly on an industrial plant to construct an accurate controller would require prohibitive amounts of data due to the random initialization of the policy. As such the initial training phase would require a model of the process dynamics, which could either be a data-driven or based on first principles. Additionally, the policy may be warm-started by techniques such as behavioral cloning or apprentice learning \cite{ALMM21}}. This would also ensure that safety violations do not occur in the real plant. The simplified workflow shown in Fig. \ref{RL_scheme} starts with either a randomly initialized policy or a policy that is warm-started by an existing controller and apprenticeship learning \cite{abbeel2004apprenticeship}. Preliminary training is performed using closed-loop simulations from the offline process model (notice that a stochastic model can be used). Here, the resulting control policy is a good approximation of the optimal policy of the real plant, which is subsequently deployed in the real plant for further training online. Importantly, system stochasticity is accounted for and the controller will continue to adapt and learn to better control and optimize the process, hence addressing plant-model mismatch \cite{del2020modifier,wang2019incremental,petsagkourakis2020safe}.
 {It is thought a number of processes with uncertain, nonlinear dynamics would particularly benefit from the maturation of RL-based controllers \cite{shin2019reinforcement}. These include biochemical reaction systems \cite{petsagkourakis2020reinforcement}, multi-phase separations  as well as those with complex rheological behaviours. However, it should be emphasised that essentially any process common to industry could benefit.}

\subsection{Stochastic Optimal Control Problem}
We assume that the stochastic dynamic system in question follows a Markov process and transitions are given by
\begin{equation}
\mathbf{x}_{t+1} \sim p\left(\mathbf{x}_{t+1} \mid \mathbf{x}_{t}, \mathbf{u}_{t}\right),\label{eq:MDP}
\end{equation}
where $p(\mathbf{x}_{t+1})$ is the probability density function of future state $\mathbf{x}_{t+1}$ given a current state $\mathbf{x}_{t} \in \mathbb{R}^{n_{x}}$ and control $\mathbf{u}_{t} \in \mathbb{R}^{n_{u}}$ at discrete time $t$, and the initial state is given by $\mathbf{x}_{0} \sim p_{\mathbf{x}_{0}}(\cdot)$. 
Without loss of generality we can write Eq. (\ref{eq:MDP}) as:
\begin{equation}
\mathbf{x}_{t+1}=f\left(\mathbf{x}_{t}, \mathbf{u}_{t}, \mathbf{d}_{t}, \mathbf{p}\right),
\end{equation}
where $\mathbf{p}\in \mathbb{R}^{n_{p}}$ are the uncertain parameters of the system and $\mathbf{d}_t \in \mathbb{R}^{n_{d}}$ are the stochastic disturbances. In this work, the goal is to maximize a predefined economic metric via a {policy $\pi$ (a function that outputs a control $\textbf{u}_t$ given state $\textbf{x}_t$)} subject to constraints. Consequently, this problem can be framed as an optimal control problem:
\begin{equation}
\mathcal{P}(\pi(\cdot)):=
\begin{dcases*}
\begin{array}{l}
\max _{\pi(\cdot)} \mathbb{E}\left\{J\left(\mathbf{x}_{0}, \ldots, \mathbf{x}_{t_f}, \mathbf{u}_{0}, \ldots, \mathbf{u}_{t_f}\right)\right\} \\
\text {s.t.} \\
\mathbf{x}_{0} \sim p_{\mathbf{x}_0}\left(\mathbf{x}_{0}\right) \\
\mathbf{x}_{t+1} \sim p\left(\mathbf{x}_{t+1} \mid \mathbf{x}_{t}, \mathbf{u}_{t}\right) \\
\mathbf{u}_{t} =\pi\left(\mathbf{x}_{t}\right) \\
\mathbf{u}_{t} \in \mathbb{U} \\
\mathbb{P}\left(\bigcap_{t=0}^{t_f}\left\{\mathbf{x}_{t} \in \mathbb{X}_{t}\right\}\right) \geq 1 - \omega \\
\forall t \in\{0, \ldots, t_f\}
\end{array}
\end{dcases*}
\label{policy}
\end{equation}
where $J$ is the objective function, $\mathbb{U}$ is the set of hard constraints for the controls and $\mathbb{X}_{t}$ denotes constraints for states that must be satisfied. In other words, 
\begin{equation}
\mathbb{X}_{t}=\left\{\mathbf{x}_{t} \in \mathbb{R}^{n_{x}} \mid g_{j, t}\left(\mathbf{x}_{t}\right) \leq 0, j=1, \ldots, n_{g}\right\},
\label{feasible_set}
\end{equation}
with $n_g$ being the total number of constraints to be satisfied, and $g_{j,t}$ being the $j$th constraint to be satisfied at time $t$. Joint constraint satisfaction must occur with high probability $1 - \omega$ where $\omega \in [0,1]$. Herein, we present a Q-learning algorithm that allows to obtain the optimal policy which satisfies joint chance constraints.

\subsection{Q-learning}
Q-learning is a model-free reinforcement learning algorithm which trains an agent to behave optimally in a Markov process \cite{watkins1992q}. The agent performs actions to maximize the expected sum of all future discounted rewards given an objective function $J(\cdot)$, which can be defined as
\begin{equation}
J=\sum_{t=0}^{t_f} \gamma^{t} R_{t}\left(\mathbf{x}_{t}, \mathbf{u}_{t}\right),
\end{equation}
where $ \gamma \in[0,1]$ is the discount factor and $R_t$ represents the reward at time $t$ given values $\mathbf{x}_t$ and $\mathbf{u}_t$. In the context of process control, the agent is the controller, which uses a policy $\pi(\cdot)$ to maximize the expected future reward through a feedback loop. Interaction between the agent (or controller) and system (in this case, a simulator) at each sampling time returns a value for the reward $R$ that represents the performance of the policy.

In Q-learning, for a policy $\pi$ an action-value function can be defined as
\begin{equation}
Q^{\pi}(\mathbf{x}_t, \mathbf{u}_t)={R}_{t+1}+\gamma \max_{\mathbf{u}_{t+1}} Q^{\pi}\left(\mathbf{x}_{t+1}, \mathbf{u}_{t+1}\right)
\label{Bellman}
\end{equation}
with $Q^{\pi}\left(\mathbf{x}_{t+1}, \mathbf{u}_{t+1}\right)$ being the expected sum of all the future rewards the agent receives in the resultant state $\mathbf{x}_{t+1}$. Importantly, the $Q$-value is the expected discounted reward for a given state and action, and therefore the optimal policy $\pi^*$ can be found using iterative updates with the Bellman equation (Eq. (\ref{Bellman})). Upon convergence, the optimal $Q$-value $Q^*$ is defined as:
\begin{equation}
Q^{*}\left(\mathbf{x}_t, \mathbf{u}_t\right)=\mathbb{E}_{\mathbf{x}_{t+1} \sim p}\left[R_{t+1}+\gamma \max _{\mathbf{u}_{t+1}} Q^{*}\left(\mathbf{x}_{t+1}, \mathbf{u}_{t+1}\right) \bigm| \mathbf{x}_t, \mathbf{u}_t\right]
\end{equation}
$Q\left(\mathbf{x}_t, \mathbf{u}_t\right)$ can be represented by function approximators such as neural networks, Gaussian process \cite{chowdhary2014off}, tree-based regressors \cite{pyeatt2001decision} amongst others.
In this work, the Q-function is approximated with a deep Q-network (DQN) $Q_\theta$ parameterized by weights $\theta$ \cite{mnih2013playing}. Here, the inputs specifically include the state $\mathbf{x}_t$, the corresponding time step $t$ and control $\mathbf{u}_t$. The DQN is trained with the use of a replay buffer that addresses the issue of correlated sequential samples \cite{lin1993reinforcement}. Huber loss is used as the error function \cite{huber1964robust, mnih2015human}. Initial exploration is encouraged using an $\epsilon$-greedy policy starting with high $\epsilon$ values, which is decayed over the course of training to ensure eventual exploitation and convergence to the optimal policy.


\section{Methodology}

\begin{figure}[h!]
    \centering
    \subfigure(a){%
        \includegraphics[scale=0.6]{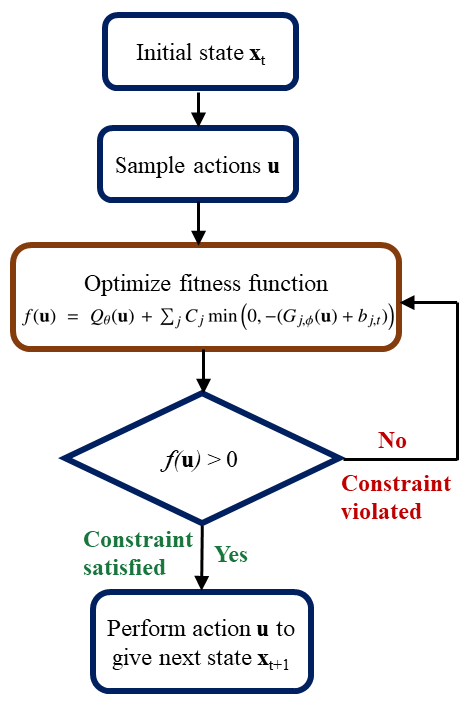}}
    \subfigure(b){%
        \includegraphics[scale=0.6]{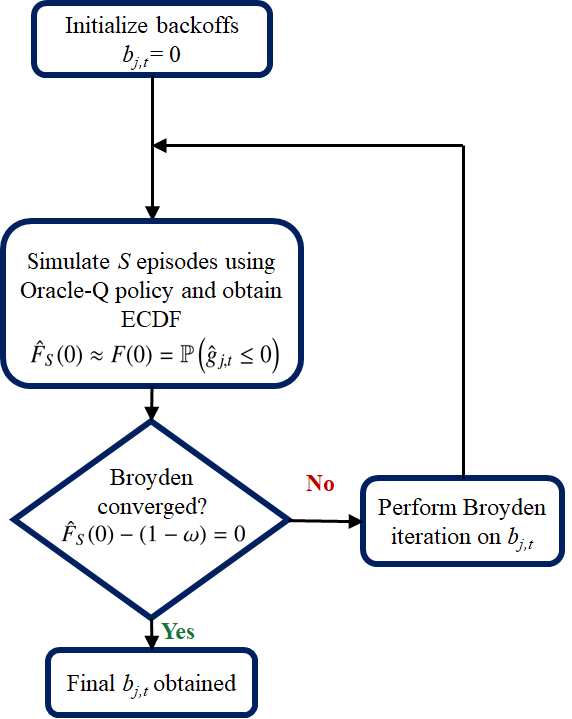}}
    \caption{Flow chart for (a) choosing action $\textbf{u}$ to satisfy constraints while maximizing reward (b) adjustment of backoffs.}
    \label{flowcharts}
\end{figure}

Our proposed approach can be found in Algorithm \ref{Oracle-assisted constrained Q-learning}. {Firstly, the Q-network ($Q_\theta$) and the constraint networks ($G_{j,\phi}$) and their respective buffers were initialized (Steps 1 and 2). Here, each constraint network $G_{j,\phi}$ corresponds to a neural network that learns the oracle values for the $j$th constraint. Subsequently, MC roll-outs of the process (Step A) generated $Q$-values (Step b) and oracle values $\hat{g}_{j,t}$ (Step c), which were stored in the respective buffers $\mathcal{D}$ and $\mathcal{G}_{j}$.} {The oracle $\hat{g}_{j,t}$ (Step c) is defined as the maximum level of violation to occur in all current and future time steps in the process realization as shown in Eq. (\ref{oracle}),} and will be further discussed in Section 3.1. {In the MC roll-outs (Step a), actions are chosen as shown in Fig. \ref{flowcharts} (a), by optimizing a fitness function (sub-problem in Algorithm \ref{Oracle-assisted constrained Q-learning}) that comprises predictions by these neural networks to ensure that chosen actions satisfy constraints while maximizing reward.} {This can be framed as the following constrained optimization problem:}

 {
\begin{equation}
\begin{array}{l}
\displaystyle\max _{\mathbf{u}} Q_{\theta}\left(\mathbf{x}, \mathbf{u}\right) \\
\text{s.t.} \\
G_{j,\phi}\left(\mathbf{x}, \mathbf{u}\right) + b_{j,t} \leq 0, j = 1, \ldots, n_g
\end{array}
\end{equation}
}

 {where $b_{j,t}$ are the backoffs which tighten the former feasible set $\mathbb{X}_{t}$ stated in Eq. (\ref{feasible_set}). The nominal case with backoff $b_{j,t} = 0$ corresponds to the absence of tightening. This is further explained in Section \ref{sec:Con tight}}
 {Neural networks ($Q_\theta$ and $G_{j,\phi}$) have been trained by random sampling from the replay/constraint buffers (Steps B and C in Algorithm \ref{Oracle-assisted constrained Q-learning}), followed by performing a gradient optimization to learn weights $\theta$ and $\phi$ (Step D).} {After training using Algorithm \ref{Oracle-assisted constrained Q-learning}, the optimal Q-network $Q_{\theta}^*$ and constraint networks $G_{j,\phi}$ are obtained.}

Subsequently, we perform constraint tightening using backoffs as described in Fig. \ref{flowcharts} (b) such that constraint satisfaction occurs with desired probability of $1-\omega$ as in Eq. (\ref{policy}), which will be discussed in Section 3.2.

\begin{algorithm}[H]

1. Initialize replay buffer $\mathcal{D}$ of size $s_\mathcal{D}$ and constraint buffers  $\mathcal{G}_{j}$ of size $s_\mathcal{{G}}$, $j = 1, \ldots, n_g$ \\
2. Initialize Q-network $Q_{\theta}$ and constraint networks $G_{j,\phi}$ with random weights $\theta$ and $\phi$, $j = 1, \ldots, n_g$ \\
3. Initialize $\epsilon$ and backoffs $b_{j,t}$ \\
4. \For {training iteration = 1, \ldots, $M$} 
    {
    A. \For {episode = 1, \ldots, N} {
        Initialize state $\mathbf{x}_0 \sim p_{\mathbf{x}_0}\left(\mathbf{x}_{0}\right)$ and episode $\mathcal{E}$
        
        a. \For{t = 0, \ldots, $t_f$}{
            With probability $\epsilon$ select random control $\mathbf{u}_{t}$
            
            otherwise select $\mathbf{u}_{t}=\max _{\mathbf{u}} Q_{\theta}\left(\mathbf{x}_{t}, \mathbf{u}_{t}\right) \mid G_{j,\phi}\left(\mathbf{x}_{t}, \mathbf{u}_{t}\right)+b_{j,t} \leq 0, j=1, \ldots, n_{g}$ (Sub-problem\footnote{
            Sub-problem: A derivative-free algorithm (e.g. evolutionary or Bayesian optimization) is used to optimize the nonconvex constrained Q-function using fitness function
            $f(\mathbf{u}) = Q_{\theta}(\mathbf{u}) + \sum_{j}C_j \min\left(0, -(G_{j, \phi}(\mathbf{u}) + b_{j,t})\right)$
            where $g_{j,t}$ is the $j$th constraint violation at time $t$, and $b_{j,t}$ is the corresponding backoff. $C_j$ are large values to ensure large negative fitness values for controls that lead to constraint violation.
        })
            
            Execute control $\mathbf{u}_{t}$ and observe reward $R_t$ and new state $\mathbf{x}_{t+1}$
            
            
            Store transition ($\mathbf{x}_{t}$, $\mathbf{u}_{t}$, $R_t$, $\mathbf{x}_{t+1}$) in $\mathcal{E}$
        }
    
        b. Extract Q-values from $\mathcal{E}$ and store datapoint ($\mathbf{x}_{t}$, $\mathbf{u}_{t}$, $Q_{t}$) in $\mathcal{D}$
        
        c. Extract oracle-constraint values from $\mathcal{E}$ using: $\hat{g}_{j, t} = \text{max}_{t' \geq t}\left[g_{j, t'}\right], j = 1, \ldots, n_g$
        
    
        d. Store datapoint ($\mathbf{x}_{t}$, $\mathbf{u}_{t}$, $\hat{g}_{j, t}$) in $\mathcal{G}_{j}, j = 1, \ldots, n_g$

    }
    B. Sample random minibatch of datapoints  of size $G$ ($\mathbf{x}_{t}$, $\mathbf{u}_{t}$, $Q_{t}$) from $\mathcal{D}$
    
    
    C. Sample random minibatch of datapoints of size $H_j$ ($\mathbf{x}_{t}$, $\mathbf{u}_{t}$, $\hat{g}_{j, t}$) from $\mathcal{G}_j$
    
    
    D. Perform a gradient-descent type step (e.g. Adam optimizer\footnote{Any other partial, or full optimization step can be used here.}) on $Q_{\theta}$ and $G_{j,\phi}$ and update weights $\theta$ and $\phi$  
    
    E. Decay $\epsilon$ using $\epsilon = D_1 \epsilon$
    
    F. Decay backoffs using $b_{j,t} = D_2 b_{j,t}$
    }
\textbf{Output:} Optimal Q-network $Q_{\theta}^*$ and constraint networks
$G_{j,\phi}, j = 1, \ldots, n_g$ 

\caption{Oracle-assisted constrained Q-learning}
\label{Oracle-assisted constrained Q-learning}

\end{algorithm}


\subsection{Oracle-assisted Constrained Q-learning}
Q-learning, when unconstrained, may offer little practical utility in process optimization due to unbounded exploration by the RL agent. For instance, an unconstrained policy may often result in a thermal runaway leading to a safety hazard in the process. As such, herein constraints $g_{j, t}$ are incorporated through the use of an oracle $\hat{g}_{j, t}$ which is formulated as
\begin{equation}
    \hat{g}_{j, t} = \text{max}_{t' \geq t}\left[g_{j, t'}\right] 
    \label{oracle}
\end{equation}
with $g_{j, t}$ being the $j$th constraint to be satisfied at time $t$, and the oracle $\hat{g}_{j, t}$ is determined by the maximum level of violation to occur in all current and future time steps $t'$ in the process realization. 

The intuition behind this framework is as follows: Imagine a car (agent) accelerating towards the wall with the goal of minimizing the time it takes to reach some distance from the wall (objective) without actually crashing into the wall (constraint). Accelerating the car without foresight causes it to go so fast that it cannot brake and stop in time, causing it to crash into the wall (constraint violated). As such, there is a need for foresight to ensure constraint satisfaction.

Effectively, the framework shown in Eq. (\ref{oracle}) is akin to an oracle (or fortune-teller peeking into a crystal ball) advising the agent on the \textit{worst} (or maximum) violation that a specific action can cause in the future given the current state. These values are easily obtained using Monte-Carlo simulations of the system. Analogous to the way a Q-function gives the sum of all future rewards, the oracle provides the worst violation in all future states if a certain action is taken by the agent, hence imbuing in the agent a sense of foresight to avoid future constraint violation.  

Similar to the Q-function, constraint values are represented by neural networks $G_{j,\phi}$ with state and action as input features. However, the subtle difference between the two is that the state representation of the input for $G_{j,\phi}$ involves time-to-termination $t_f - t$ instead of time $t$ for the case of batch processes. 

\subsection{Constraint Tightening}\label{sec:Con tight}
To satisfy the constraints with high probability, constraints are tightened with backoffs \cite{bradford2020stochastic, rafiei2018stochastic} $b_{j,t}$ as:
\begin{equation}
\overline{\mathbb{X}}_{t}=\left\{\mathbf{x}_{t} \in \mathbb{R}^{n_{x}} \mid g_{j, t}\left(\mathbf{x}_{t}\right)+b_{j,t} \leq 0, j=1, \ldots, n_{g}\right\}
\end{equation}
where 
$b_{j,t}$ are the backoffs which tighten the former feasible set $\mathbb{X}_{t}$ stated in Eq. (\ref{feasible_set}). The result of this would be the reduction of the perceived feasible space by the agent, which consequently allows for the satisfaction of constraints. Notice that the value of the backoffs necessarily imply a trade-off: large backoff values ensure constraint satisfaction, but renders the policy over-conservative hence sacrificing performance. Conversely, smaller backoff values afford solutions with higher rewards, but may not guarantee constraint satisfaction. Therefore, the values of $b_{j,t}$ are the minimum value needed to guarantee satisfaction of constraints.

To determine the desired backoffs, the cumulative distribution function (CDF) $F$ of the oracle $\hat{g}_{j, t}$ is approximated using sample average approximation (SAA) with $S$ Monte Carlo (MC) simulations to give its empirical cumulative distribution function (ECDF) $\hat{F}_S$ where
\begin{equation}
\hat{F}_S(0) \approx F(0)=\mathbb{P}\left( \hat{g}_{j, t} \leq 0 \right)
\label{ECDF}
\end{equation}
hence $\hat{F}_S(0)$ is the approximate probability for a trajectory to satisfy a constraint. We can therefore pose a root-finding problem such that we adjust the backoffs $b_{j,t}$ to find:

\begin{equation}
\hat{F}_S(0) - (1 - \omega) = 0
\label{backoffs}
\end{equation}

We solve Eq. (\ref{backoffs}) via the  quasi-Newton Broyden's method \cite{kelley1995iterative} given its fast convergence near optimal solutions. Where $\omega$ is a tunable parameter depending on the case study, such that constraint satisfaction occurs with high probability $1-\omega$ as shown in Eq. (\ref{policy}). Alternatively, the empirical lower bound of the ECDF can be forced to be $1-\omega$, and guarantee with confidence $1-\epsilon$ that $\mathbb{P}\left( \hat{g}_{j, t} \leq 0 \right) \geq 1 - \omega $. Note that this constraint tightening takes place directly during the constraunction of $G_{j,\phi},j=1,...,n_g$

\section{Case Studies}
\subsection{Case Study 1}
This case study pertains to the photoproduction of phycocyanin synthesized by cyanobacterium \textit{Arthrospira platensis}. Phycocyanin is a high-value bioproduct, and serves its biological
role by increasing the photosynthetic efficiency of cyanobacteria and red algae \cite{bradford2020stochastic}. In addition, it is used as a natural colorant to substitute toxic synthetic pigments in cosmetic and food manufacturing. Moreover, it possesses antioxidant, and anti-inflammatory properties.

The dynamic system comprises a system of ODEs from \cite{bradford2020stochastic} that describes the evolution of concentration ($c$) of biomass ($x$), nitrate ($N$) and product ($q$) under parametric uncertainty. The model is based on Monod kinetics, which describes the growth of microorganism in nutrient-sufficient cultures, where intracellular nutrient concentration is kept constant because of rapid replenishment. Here, a fixed volume fed-batch is assumed. The controls are light intensity ($u_1 = I$) and inflow rate ($u_2 = F_N$). 
The mass balance equations are as follows:
\begin{equation}
\begin{aligned}
\frac{d c_{x}}{d t} &=u_{m} \frac{I}{I+k_{s}+I^{2} / k_{i}} \frac{c_{x} c_{N}}{c_{N}+K_{N}}-u_{d} c_{x} \\
\frac{d c_{N}}{d t} &=-Y_{N / X} \frac{u_{m} I}{I+k_{s}+I^{2} / k_{i}} \frac{c_{x} c_{N}}{c_{N}+K_{N}}+F_{N} \\
\frac{d c_{q}}{d t} &=\frac{k_{m} I}{I+k_{s q}+I^{2} / k_{i q}} c_{x} -\frac{k_{d} c_{q}}{c_{N}+K_{N_{q}}}
\end{aligned}
\end{equation}

This case study and parameter values are adopted from \cite{bradford2020stochastic}. Uncertainty in the system is two-fold: First, the initial concentration adopts a Gaussian distribution, where $[c_{x,0}, c_{N,0}] \sim \mathcal{N}([1.0,  150.0]$, diag$(10^{-3}, 22.5))$ and $c_q(0) = 0$. Second, parametric uncertainty is assumed to be: 
$
\frac{k_{s}}{\left(\mu m o l / m^{2} / s\right)} \sim \mathcal{N}(178.9,\sigma_{k_s}^2)$,
$\frac{k_{i}}{(m g / L)} \sim\mathcal{N}(447.1,\sigma_{k_i}^2)$,
$\frac{k_{N}}{\left(\mu m o l / m^{2} / s\right)} \sim \mathcal{N}(393.1,\sigma_{k_N}^2)$
where the variance $\sigma_{i}^2 = 10 \%$ of its corresponding mean value. This type of uncertainty is common in engineering settings, as the parameters are experimentally determined, and therefore subject to confidence intervals after being calculated through nonlinear regression techniques. The objective function is to maximize the product concentration ($c_q$) at the end of the batch, hence the reward is defined as:
\begin{equation}
R_{t_f}=c_{q, t_f}
\end{equation}
where $t_f$ is the terminal time step. 
The two path constraints are as follows: Nitrate concentration ($c_N$) is to remain below $800$ mg/L, and the ratio of bioproduct concentration ($c_q$) to biomass concentration ($c_x$) cannot exceed $11.0$ mg/g for high density biomass cultivation. These constraints can be formulated as:
\begin{equation}
\begin{aligned}
g_{1, t} = c_{N}-800 \leq 0 \quad \forall t \in\{0, \ldots, t_f\} \\
g_{2, t} = c_{q}-0.011 c_{x} \leq 0 \quad \forall t \in\{0, \ldots, t_f\} \\
\end{aligned}
\end{equation}
The control inputs are subject to hard constraints to be in the interval $0 \leq F_{N} \leq 40$ and $120 \leq I \leq 400$. The time horizon was set to 12 with an overall batch time of 240 h, and hence giving a sampling time of 20 h. The Q-network $Q_\theta$ consists of 2 fully connected hidden layers, each consisting of 200 neurons with a leaky rectified linear unit (LeakyReLU) as activation function. The parameters used in Algorithm \ref{Oracle-assisted constrained Q-learning} for training the agent are: $\epsilon = 0.99$, $b_{1,t} = -500$, $b_{2,t} = -0.05$, $s_\mathcal{D} = 3000$, $s_\mathcal{{G}} = 30000$, $M = 2000$, $N = 100$, $G = 100$, $H_1 = 500$, $H_2 = 1000$, $D_1 = 0.99$ and $D_2 = 0.995$. Upon completion of training, validation was conducted via the trained policy with respect to the Q function. The policy is optimized through an evolutionary strategy \cite{slowik2020evolutionary} given its nonconvex nature, as discussed in Algorithm \ref{Oracle-assisted constrained Q-learning} and Fig. \ref{flowcharts} (a).

\begin{figure}[h!]
    \centering
    \subfigure(a){%
        \includegraphics[scale=0.23]{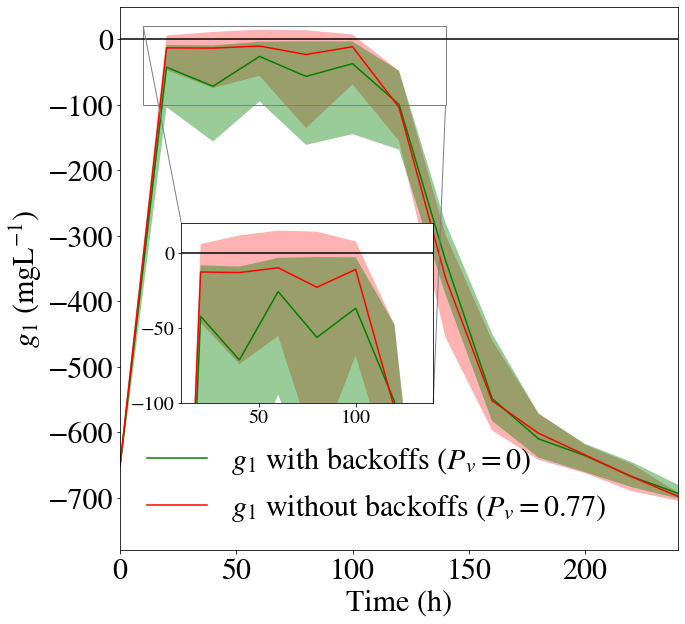}
        }
    \subfigure(b){%
        \includegraphics[scale=0.23]{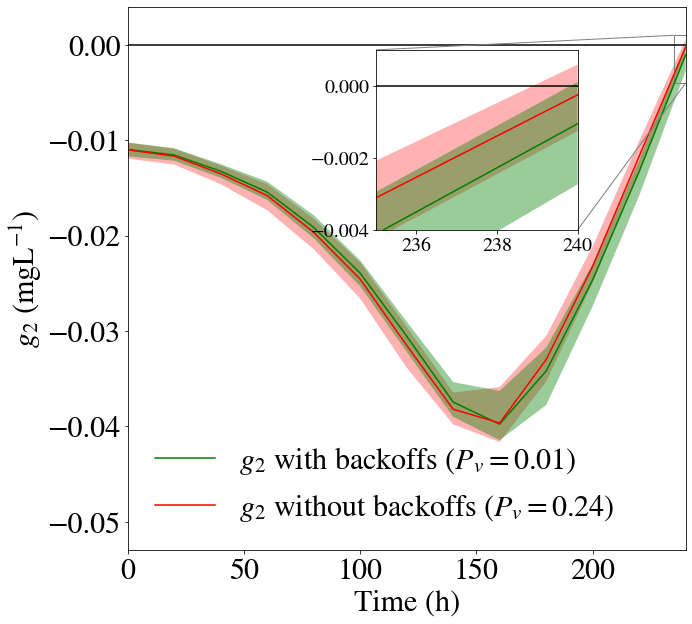}
        }
    \caption{Case Study 1: Constraints $g_{1,t}$ (a) and $g_{2,t}$ (b) when backoffs are applied (green), and when they are absent (red) with probabilities of violation $P_v$ within the parentheses. Inset: Zoomed-in region where violation of constraints occur. Solid lines represent the expected values. Shaded areas represent the 99th to 1st percentiles.}
    \label{g1g2_backoff_vs_no_backoff}
\end{figure}
\begin{figure}[h!]
    \centering
    \subfigure(a){%
        \includegraphics[scale=0.23]{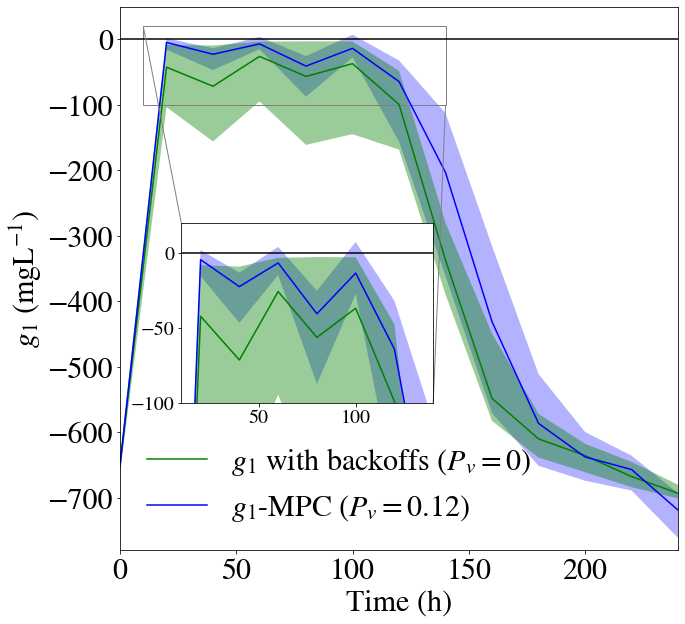}
        }
    \subfigure(b){%
        \includegraphics[scale=0.23]{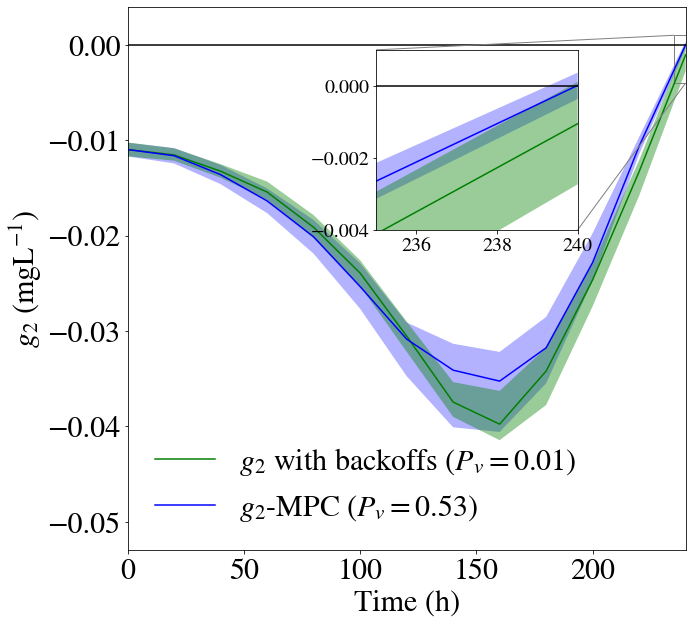}
        }
    \caption{Case Study 1: Constraints $g_{1,t}$ (a) and $g_{2,t}$ (b) when backoffs are applied (green), and for NMPC (blue) with probabilities of violation $P_v$ within the parentheses. Inset: Zoomed-in region where violation of constraints occur. Solid lines represent the expected values. Shaded areas represent the 99th to 1st percentiles.}
    \label{g1g2_backoff_vs_MPC}
\end{figure}
\begin{table}[h!]
  \caption{Case Study 1: Comparison of probabilities of constraint violation $P_v$ and objective values of different algorithms}
  \label{CS1_table}
  \centering
  \begin{tabular}{lll}
    \toprule

    Algorithm & Violation probability $P_v$ & Objective ($c_{q,t_f}$) \\
    \midrule
    Oracle Q-learning with backoffs    & \textbf{0.01} & 0.166 \\
    Oracle Q-learning without backoffs & 0.82 & 0.169  \\
    NMPC                         & 0.53 & 0.168  \\
    \bottomrule
  \end{tabular}
\end{table}
After completion of training using Algorithm \ref{Oracle-assisted constrained Q-learning}, the backoffs are adjusted to satisfy Eq. (\ref{backoffs}) with $S = 1000$ and $\omega = 0.01$, with backoffs at all time-steps $t$ being constant. 
The constraint satisfaction is shown in Fig. \ref{g1g2_backoff_vs_no_backoff}, where the shaded areas represent the 99th to 1st percentiles. Here, we elucidate the importance of applying backoffs to the policy: As shown in Fig. \ref{g1g2_backoff_vs_no_backoff} (a), even though it may seem at face value that $g_{1,t}$ values for both methods are similar, the zoomed-in region (in the inset) clearly shows that oracle Q-learning without backoffs (red) results in a high probability of constraint violation ($P_v = 0.77$), with parts of the red shaded regions exceeding zero. The violation probabilities $P_v$ in Fig. \ref{g1g2_backoff_vs_no_backoff} and \ref{g1g2_backoff_vs_MPC} correspond to the fraction of 400 MC trajectories that violate a certain constraint. Gratifying, when backoffs are applied (green) in Fig. \ref{g1g2_backoff_vs_no_backoff} (a), all constraints are satisfied ($P_v = 0$), as shown by all the green shaded regions staying below zero. 

In the same vein, in Fig. \ref{g1g2_backoff_vs_no_backoff} (b), applying backoffs resulted in a drastic reduction of constraint violation from $P_v = 0.24$ to $0.01$. This is expected since the backoffs are adjusted using $\omega$ = 0.01 in Eq. (\ref{policy}). The objective value, represented by the final concentration of product $c_q$, are 0.166 and 0.169 for oracle Q-learning with and without backoffs, respectively. Consequently, this indicates that a small compromise in objective value can result in high probability of constraint satisfaction, where violation probability is reduced from 0.82 to 0.01 (in boldface) upon applying backoffs as shown in Table \ref{CS1_table}.

In addition, the performance of the oracle Q-learning algorithm with backoffs has been compared with that of nonlinear NMPC using the nominal parameters of the model, which is one of the main process control techniques used in chemical process optimization and hence serves as an important benchmark. 

Although NMPC achieves a slightly higher objective value (Table \ref{CS1_table}), it fares poorly in terms of constraint satisfaction as shown in blue Fig. \ref{g1g2_backoff_vs_MPC} (a) and (b) where probabilities of violation are 12 and 53\% for $g_1$ and $g_2$, respectively. This is unsurprising, since NMPC is only able to satisfy constraints in \textit{expectation}, which means that in a stochastic system, loosely speaking, violation occurs 50\% of the time. On the other hand, oracle Q-learning with backoffs violated a constraint only 1 \% of the time (boldface in Table \ref{CS1_table}). Therefore, it is clear that this algorithm offers a more effective means of handling constraints compared to NMPC. 

 {It is also noteworthy to contrast the proposed method with policy-based methods. Previous work in the group \cite{petsagkourakis2020chance} involves policy optimization, which has its own benefits, such as avoiding an online optimization routine, this however, comes with the drawback of being unable to handle constraints naturally. On the other hand, value-based methods, as proposed in this paper, have an inner optimization loop, which allows them to handle constraints easily with mature algorithms from the numerical optimization community. Therefore, value-based methods can be better tailored to satisfy constraints.}

\subsection{Case Study 2}
The second case study involves a challenging semi-batch reactor adopted from \cite{bradford2018economic}, with the following chemical reactions in the reactor catalyzed by H$_2$SO$_4$:
\begin{equation}
2 \mathrm{A} 
\xrightarrow[(1)]{k_{\text{1A}}}
\mathrm{B} 
\xrightarrow[(2)]{k_{\text{2B}}}
3 \mathrm{C}
\end{equation}
Here, the reactions are first-order. Reactions (1) and (2) are exothermic and endothermic, respectively. The temperature is controlled by a cooling jacket. The controls are the flowrate of reactant A entering the reactor and the temperature of the cooling jacket $T_0$. Therefore, the state is represented by the concentrations of A, B, and C in mol/L ($c_A$, $c_B$, $c_C$), reactor temperature in K ($T$), and the reactor volume in L ($Vol$). The model of the physical system can be found in \cite{bradford2018economic}.

The objective function is to maximize the amount of product ($c_C \cdot Vol$) at the end of the batch. Two path constraints exist. Firstly, the reactor temperature needs to be below 420 K due to safety reasons and secondly, the reactor volume is required to be below the maximum reactor capacity of 800 L and therefore: 
\begin{equation}
\begin{aligned}
g_{1, t} = T - 420 \leq 0 \quad \forall t \in\{0, \ldots, t_f\} \\
g_{2, t} = Vol - 800 \leq 0 \quad \forall t \in\{0, \ldots, t_f\} \\
\end{aligned}
\end{equation}
The time horizon is fixed to 10 with an overall batch time of 4 h, therefore the sampling time is 0.4 h. Parametric uncertainty is set as: 
$
\theta_{1} \sim \mathcal{N}(4,0.1)$, 
$ A_2 \sim\mathcal{N}(0.08,1.6 \times10^{-4})$, 
$ \theta_4 \sim \mathcal{N}(100, 5)$. The initial concentrations of A, B and C are set to zero. The initial reactor temperature and volume are 290 K and 100 L, respectively.

\begin{figure}[h!]
    \centering
    \subfigure(a){%
        \includegraphics[scale=0.23]{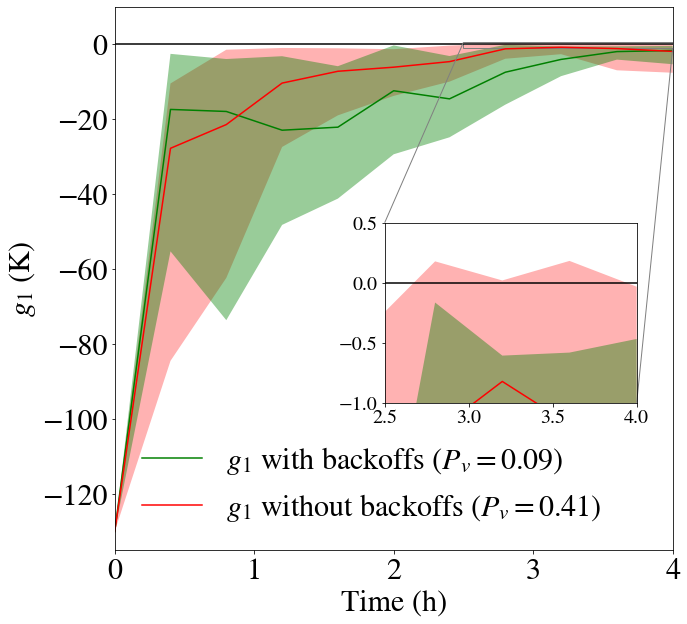}
        }
    \subfigure(b){%
        \includegraphics[scale=0.23]{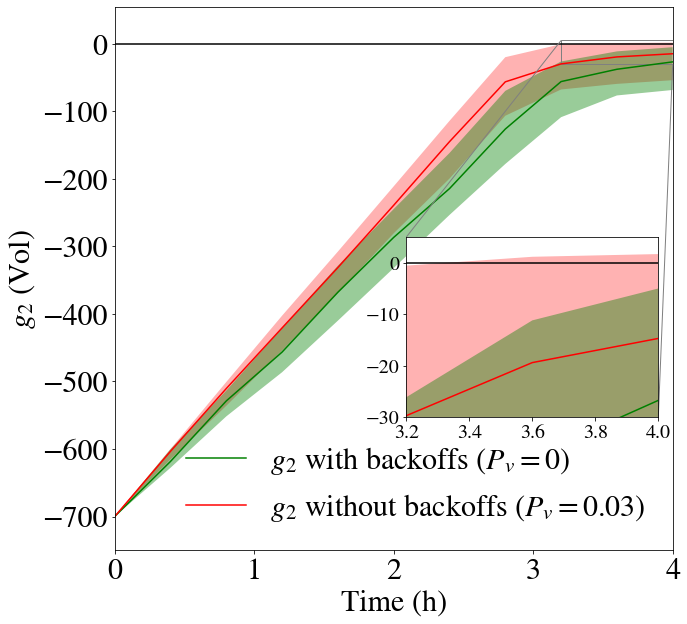}
        }
    \caption{Case Study 2: Constraints $g_{1,t}$ (a) and $g_{2,t}$ (b) when backoffs are applied (green), and when they are absent (red) with probabilities of violation $P_v$ within the parentheses. Inset: Zoomed-in region where violation of constraints occur. Solid lines represent the expected values. Shaded areas represent the 95th-5th percentiles for (a) and 99th-1st percentiles for (b).}
    \label{CS2_g1g2_backoff_vs_no_backoff}
\end{figure}

In this case study, due to its more challenging nature in terms of constraint satisfaction compared to the first case study, the backoffs have been adjusted to satisfy Eq. (\ref{backoffs}) using $\omega = 0.1$ in Eq. (\ref{policy}). We observe that backoffs again proved to be necessary to ensure high probability of constraint satisfaction. From the inset of Fig. \ref{CS2_g1g2_backoff_vs_no_backoff} (a), we can see that without backoffs the policy violates $g_1$ 41\% of the time, and this probability is reduced to 9\% when backoffs are applied. The same applies for $g_2$ in Fig. \ref{CS2_g1g2_backoff_vs_no_backoff} (b) where $P_v$ is completely eliminated from 3\% to 0\% using backoffs.

\begin{figure}[h!]
    \centering
    \subfigure(a){%
        \includegraphics[scale=0.23]{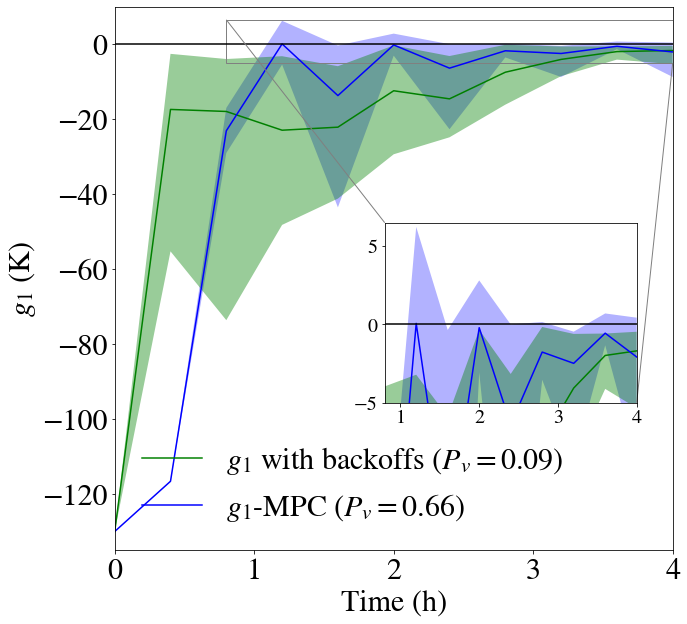}
        }
    \subfigure(b){%
        \includegraphics[scale=0.23]{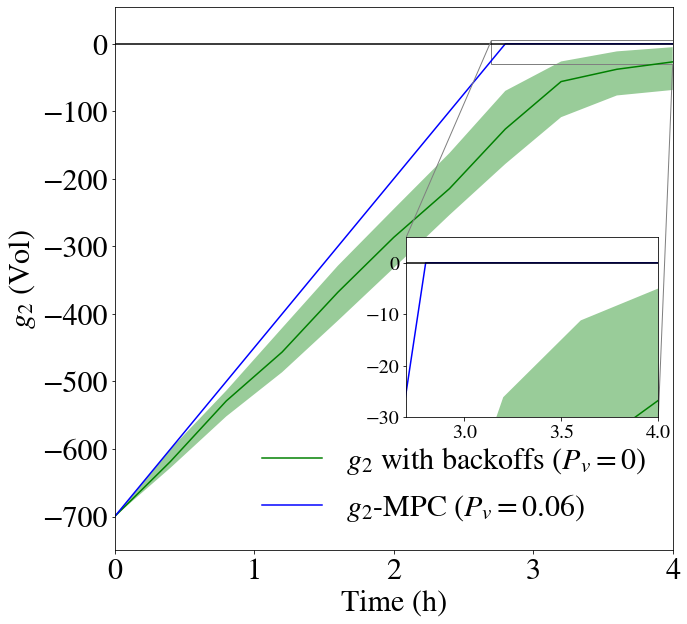}
        }
    \caption{Case Study 2: Constraints $g_{1,t}$ (a) and $g_{2,t}$ (b) when backoffs are applied (green), and for NMPC (blue) with probabilities of violation $P_v$ within the parentheses. Inset: Zoomed-in region where violation of constraints occur. Solid lines represent the expected values. Shaded areas represent the 95th-5th percentiles for (a) and 99th-1st percentiles for (b).}
    \label{CS2_g1g2_backoff_vs_MPC}
\end{figure}

\begin{table}[h!]
  \caption{Case Study 2: Comparison of probabilities of constraint violation $P_v$ and objective values of different algorithms}
  \label{CS2_table}
  \centering
  \begin{tabular}{lll}
    \toprule

    Algorithm & Violation probability $(P_v)$ & Objective ($c_{C,t_f} \cdot Vol_{t_f}$)\\
    \midrule
    Oracle Q-learning with backoffs    & \textbf{0.09} & 532 \\
    Oracle Q-learning without backoffs & 0.44 & 680  \\
    NMPC                         & 0.66 & 714  \\
    \bottomrule
  \end{tabular}
\end{table}
 {
To compare the performance of NMPC with oracle Q-learning with backoffs in the context of this case study. The average runtime to solve for an individual control for the oracle Q-learning approach (0.03 $\pm$ 0.01 s) is faster than NMPC (0.4 $\pm$ 0.1 s). This can be attributed to the fact that MPC relies on a model-based optimization, whereas our approach has already learnt this offline.}
Moreover, it is noteworthy that chemical systems are rarely deterministic in nature, hence limiting the applicability of NMPC. In a stochastic system, NMPC often struggles in terms of constraint handling. This can be clearly seen in Fig. \ref{CS2_g1g2_backoff_vs_MPC} (a), where the MPC trajectories only satisfy $g_1$ in expectation (blue line), hence resulting in high levels of violations (66\%). Intriguingly, for $g_2$ the NMPC trajectory in Fig. \ref{CS2_g1g2_backoff_vs_MPC} displayed little variation, resulting in only small probability of violation (6\%). 

In terms of objective values, unlike the first case study, oracle Q-learning with backoffs saw a significant decrease in objective value in Table \ref{CS2_table} after applying backoffs. This is expected because we further restrict the feasible space of the controller leading to a more conservative solution, hence exhibiting a trade-off between constraint satisfaction and objective value. 

This trade-off is justified as the NMPC solution results in 66\% probability of constraint violation (Table \ref{CS2_table}). In the context of a chemical plant, the NMPC solution is infeasible due to the high risk. Process operation in such industries necessitates that these probabilities are minimized as safety is of utmost importance in chemical engineering. This provides basis for the use of RL in the process industries.

On the other hand, for oracle Q-learning, it can been seen that the probability of constraint violation has been significantly improved from 66\% (for NMPC) to 9\% (boldface in Table \ref{CS2_table}). Clearly, oracle Q-learning offers an effective means of not only satisfying constraints in expectation (green lines in Fig. \ref{CS2_g1g2_backoff_vs_no_backoff}), but more importantly with high probability (all green shaded areas below zero).

However, it is worth noting that this algorithm is based on Q-learning, which is expected to take longer time to train than MPC, particularly because it requires backoffs to be tuned. This is a direct consequence of shifting the computation time from online to offline. Indeed, such a tradeoff can be justified as this guarantees robust constraint satisfaction \textit{online} with fast computation time, which is crucial in many safety critical engineering applications.


\section{Conclusions}
In this paper we propose a new reinforcement learning methodology for finding a controller policy that can satisfy constraints with high probability in stochastic and complex process systems. The proposed algorithm - oracle-assisted constrained Q-learning - uses constraint tightening by applying backoffs to the original feasible set. Backoffs restrict the perceived feasible space by the controller, hence allowing guarantees on the satisfaction of chance constraints. Here, we find the smallest backoffs (least
conservative) that still guarantee the desired probability of satisfaction by solving a root-finding problem using Broyden's method. Results show that our proposed methodology compares favorably to nonlinear model predictive control (NMPC), a benchmark control technique commonly used in the industry, in terms of constraint handling. This is expected since NMPC guarantees constraint satisfaction only in \textit{expectation} (loosely speaking constraints are satisfied only 50\% of the time), while our algorithm ensures constraint satisfaction with probabilities as high as 99\% as shown in the case studies. Being able to solve constraint policy optimization problems with high probability constraint satisfaction has been one of the main hurdles of the widespread use of RL in engineering applications. The promising performance of this algorithm is an encouraging step towards applying RL to real-world industrial chemical processes, where constraints on policies are absolutely critical due to safety reasons.
\section*{Acknowledgment}
This project has received funding from the EPSRC projects (EP/R032807/1) and (EP/P016650/1).
\bibliography{references}

\begin{thebibliography}{60}
\expandafter\ifx\csname natexlab\endcsname\relax\def\natexlab#1{#1}\fi
\expandafter\ifx\csname url\endcsname\relax
  \def\url#1{\texttt{#1}}\fi
\expandafter\ifx\csname urlprefix\endcsname\relax\def\urlprefix{URL }\fi

\bibitem[{Abbeel and Ng(2004)}]{abbeel2004apprenticeship}
P.~Abbeel, A.~Y. Ng, 2004. Apprenticeship learning via inverse reinforcement
  learning. In: Proceedings of the twenty-first international conference on
  Machine learning. p.~1.

\bibitem[{Achiam et~al.(2017)Achiam, Held, Tamar, and
  Abbeel}]{achiam2017constrained}
J.~Achiam, D.~Held, A.~Tamar, P.~Abbeel, 2017. Constrained policy optimization.

\bibitem[{Altman(1999)}]{altman1999constrained}
E.~Altman, 1999. Constrained Markov decision processes. Vol.~7. CRC Press.

\bibitem[{Bertsekas et~al.(1995)Bertsekas, Bertsekas, Bertsekas, and
  Bertsekas}]{bertsekas1995dynamic}
D.~P. Bertsekas, D.~P. Bertsekas, D.~P. Bertsekas, D.~P. Bertsekas, 1995.
  Dynamic programming and optimal control. Vol.~1. Athena scientific Belmont,
  MA.

\bibitem[{Bradford and Imsland(2018)}]{bradford2018economic}
E.~Bradford, L.~Imsland, 2018. Economic stochastic model predictive control
  using the unscented kalman filter. IFAC-PapersOnLine 51~(18), 417--422.

\bibitem[{Bradford et~al.(2020)Bradford, Imsland, Zhang, and del
  Rio~Chanona}]{bradford2020stochastic}
E.~Bradford, L.~Imsland, D.~Zhang, E.~A. del Rio~Chanona, 2020. Stochastic
  data-driven model predictive control using gaussian processes. Computers \&
  Chemical Engineering 139, 106844.

\bibitem[{Buckman(2021)}]{RLstreamorsave}
J.~Buckman, 2021. How to think about replay memory.
\newline\urlprefix\url{https://jacobbuckman.com/2021-02-13-how-to-think-about-replay-memory/}

\bibitem[{Cheng et~al.(2019)Cheng, Orosz, Murray, and Burdick}]{cheng2019end}
R.~Cheng, G.~Orosz, R.~M. Murray, J.~W. Burdick, 2019. End-to-end safe
  reinforcement learning through barrier functions for safety-critical
  continuous control tasks. In: Proceedings of the AAAI Conference on
  Artificial Intelligence. Vol.~33. pp. 3387--3395.

\bibitem[{Choi et~al.(2020)Choi, Casta{\~n}eda, Tomlin, and
  Sreenath}]{choi2020reinforcement}
J.~Choi, F.~Casta{\~n}eda, C.~J. Tomlin, K.~Sreenath, 2020. Reinforcement
  learning for safety-critical control under model uncertainty, using control
  lyapunov functions and control barrier functions. arXiv preprint
  arXiv:2004.07584.

\bibitem[{Chow et~al.(2019)Chow, Nachum, Faust, Duenez-Guzman, and
  Ghavamzadeh}]{Chow2019}
Y.~Chow, O.~Nachum, A.~Faust, E.~Duenez-Guzman, M.~Ghavamzadeh, 2019.
  {Lyapunov-based safe policy optimization for continuous control}.
\newline\urlprefix\url{http://arxiv.org/abs/1901.10031}

\bibitem[{Chowdhary et~al.(2014)Chowdhary, Liu, Grande, Walsh, How, and
  Carin}]{chowdhary2014off}
G.~Chowdhary, M.~Liu, R.~Grande, T.~Walsh, J.~How, L.~Carin, 2014. Off-policy
  reinforcement learning with gaussian processes. IEEE/CAA Journal of
  Automatica Sinica 1~(3), 227--238.

\bibitem[{Dalal et~al.(2018)Dalal, Dvijotham, Vecer{\'{\i}}k, Hester, Paduraru,
  and Tassa}]{Safefilter}
G.~Dalal, K.~Dvijotham, M.~Vecer{\'{\i}}k, T.~Hester, C.~Paduraru, Y.~Tassa,
  2018. Safe exploration in continuous action spaces. CoRR abs/1801.08757.
\newline\urlprefix\url{http://arxiv.org/abs/1801.08757}

\bibitem[{del Rio~Chanona et~al.(2021)del Rio~Chanona, Petsagkourakis,
  Bradford, Graciano, and Chachuat}]{del2020modifier}
E.~A. del Rio~Chanona, P.~Petsagkourakis, E.~Bradford, J.~E.~A. Graciano,
  B.~Chachuat, 2021. Real-time optimization meets bayesian optimization and
  derivative-free optimization: A tale of modifier adaptation. Computers \&
  Chemical Engineering 147, 107249.

\bibitem[{Engstrom et~al.(2020)Engstrom, Ilyas, Santurkar, Tsipras, Janoos,
  Rudolph, and Madry}]{engstrom2020implementation}
L.~Engstrom, A.~Ilyas, S.~Santurkar, D.~Tsipras, F.~Janoos, L.~Rudolph,
  A.~Madry, 2020. Implementation matters in deep policy gradients: A case study
  on ppo and trpo. arXiv preprint arXiv:2005.12729.

\bibitem[{Goulart and Pereira(2020)}]{goulart2020nautonomous}
D.~A. Goulart, R.~D. Pereira, 2020. Autonomous ph control by reinforcement
  learning for electroplating industry wastewater. Computers \& Chemical
  Engineering, 106909.

\bibitem[{Haarnoja et~al.(2018)Haarnoja, Zhou, Abbeel, and
  Levine}]{haarnoja2018soft}
T.~Haarnoja, A.~Zhou, P.~Abbeel, S.~Levine, 2018. Soft actor-critic: Off-policy
  maximum entropy deep reinforcement learning with a stochastic actor. arXiv
  preprint arXiv:1801.01290.

\bibitem[{Huber(1964)}]{huber1964robust}
P.~J. Huber, 1964. Robust estimation of a location parameter: Annals
  mathematics statistics, 35.

\bibitem[{Huh and Yang(2020)}]{huh2020safe}
S.~Huh, I.~Yang, 2020. Safe reinforcement learning for probabilistic
  reachability and safety specifications: A lyapunov-based approach. arXiv
  preprint arXiv:2002.10126.

\bibitem[{Hwangbo and Sin(2020)}]{hwangbo2020design}
S.~Hwangbo, G.~Sin, 2020. Design of control framework based on deep
  reinforcement learning and monte-carlo sampling in downstream separation.
  Computers \& Chemical Engineering, 106910.

\bibitem[{Kelley(1995)}]{kelley1995iterative}
C.~T. Kelley, 1995. Iterative methods for linear and nonlinear equations. SIAM.

\bibitem[{Lawrence et~al.(2020)Lawrence, Stewart, Loewen, Forbes, Backstrom,
  and Gopaluni}]{lawrence2020optimal}
N.~P. Lawrence, G.~E. Stewart, P.~D. Loewen, M.~G. Forbes, J.~U. Backstrom,
  R.~B. Gopaluni, 2020. Optimal pid and antiwindup control design as a
  reinforcement learning problem. arXiv preprint arXiv:2005.04539.

\bibitem[{Lee and Lee(2006)}]{lee2006approximate}
J.~H. Lee, J.~M. Lee, 2006. Approximate dynamic programming based approach to
  process control and scheduling. Computers \& chemical engineering 30~(10-12),
  1603--1618.

\bibitem[{Lee and Lee(2005)}]{lee2005approximate}
J.~M. Lee, J.~H. Lee, 2005. Approximate dynamic programming-based approaches
  for input-output data-driven control of nonlinear processes. Automatica
  41~(7), 1281--1288.

\bibitem[{Lehman et~al.(2018)Lehman, Chen, Clune, and Stanley}]{lehman2018more}
J.~Lehman, J.~Chen, J.~Clune, K.~O. Stanley, 2018. Es is more than just a
  traditional finite-difference approximator. In: Proceedings of the Genetic
  and Evolutionary Computation Conference. pp. 450--457.

\bibitem[{Leurent et~al.(2020)Leurent, Efimov, and
  Maillard}]{leurent2020robustadaptive}
E.~Leurent, D.~Efimov, O.-A. Maillard, 2020. Robust-adaptive control of linear
  systems: beyond quadratic costs.

\bibitem[{Lillicrap et~al.(2015)Lillicrap, Hunt, Pritzel, Heess, Erez, Tassa,
  Silver, and Wierstra}]{lillicrap2015continuous}
T.~P. Lillicrap, J.~J. Hunt, A.~Pritzel, N.~Heess, T.~Erez, Y.~Tassa,
  D.~Silver, D.~Wierstra, 2015. Continuous control with deep reinforcement
  learning. arXiv preprint arXiv:1509.02971.

\bibitem[{Lin(1993)}]{lin1993reinforcement}
L.-J. Lin, 1993. Reinforcement learning for robots using neural networks. Tech.
  rep., Carnegie-Mellon Univ Pittsburgh PA School of Computer Science.

\bibitem[{Mehta and Ricardez-Sandoval(2016)}]{mehta2016integration}
S.~Mehta, L.~A. Ricardez-Sandoval, 2016. Integration of design and control of
  dynamic systems under uncertainty: A new back-off approach. Industrial \&
  Engineering Chemistry Research 55~(2), 485--498.

\bibitem[{Mnih et~al.(2013)Mnih, Kavukcuoglu, Silver, Graves, Antonoglou,
  Wierstra, and Riedmiller}]{mnih2013playing}
V.~Mnih, K.~Kavukcuoglu, D.~Silver, A.~Graves, I.~Antonoglou, D.~Wierstra,
  M.~Riedmiller, 2013. Playing atari with deep reinforcement learning. arXiv
  preprint arXiv:1312.5602.

\bibitem[{Mnih et~al.(2015)Mnih, Kavukcuoglu, Silver, Rusu, Veness, Bellemare,
  Graves, Riedmiller, Fidjeland, Ostrovski, et~al.}]{mnih2015human}
V.~Mnih, K.~Kavukcuoglu, D.~Silver, A.~A. Rusu, J.~Veness, M.~G. Bellemare,
  A.~Graves, M.~Riedmiller, A.~K. Fidjeland, G.~Ostrovski, et~al., 2015.
  Human-level control through deep reinforcement learning. nature 518~(7540),
  529--533.

\bibitem[{Mowbray et~al.(2021)Mowbray, Smith, Del Rio-Chanona, and
  Zhang}]{ALMM21}
M.~Mowbray, R.~Smith, E.~A. Del Rio-Chanona, D.~Zhang, 2021. Using process data
  to generate an optimal control policy via apprenticeship and reinforcement
  learning. Submitted to Journal.

\bibitem[{Ng et~al.(1999)Ng, Harada, and Russell}]{ng1999policy}
A.~Y. Ng, D.~Harada, S.~Russell, 1999. Policy invariance under reward
  transformations: Theory and application to reward shaping. In: ICML. Vol.~99.
  pp. 278--287.

\bibitem[{Nocedal and Wright(2006)}]{nocedal2006numerical}
J.~Nocedal, S.~Wright, 2006. Numerical optimization. Springer Science \&
  Business Media.

\bibitem[{Peng et~al.(2021)Peng, Mu, Duan, Guan, Li, and
  Chen}]{peng2021separated}
B.~Peng, Y.~Mu, J.~Duan, Y.~Guan, S.~E. Li, J.~Chen, 2021. Separated
  proportional-integral lagrangian for chance constrained reinforcement
  learning.

\bibitem[{Petsagkourakis and Galvanin(2020)}]{petsagkourakis2020safe}
P.~Petsagkourakis, F.~Galvanin, 2020. Safe model-based design of experiments
  using gaussian processes.

\bibitem[{Petsagkourakis et~al.(2020{\natexlab{a}})Petsagkourakis, Sandoval,
  Bradford, Galvanin, Zhang, and del Rio-Chanona}]{petsagkourakis2020chance}
P.~Petsagkourakis, I.~O. Sandoval, E.~Bradford, F.~Galvanin, D.~Zhang, E.~A.
  del Rio-Chanona, 2020{\natexlab{a}}. Chance constrained policy optimization
  for process control and optimization. arXiv preprint arXiv:2008.00030.

\bibitem[{Petsagkourakis et~al.(2020{\natexlab{b}})Petsagkourakis, Sandoval,
  Bradford, Zhang, and del Rio-Chanona}]{petsagkourakis2020reinforcement}
P.~Petsagkourakis, I.~O. Sandoval, E.~Bradford, D.~Zhang, E.~A. del
  Rio-Chanona, 2020{\natexlab{b}}. Reinforcement learning for batch bioprocess
  optimization. Computers \& Chemical Engineering 133, 106649.

\bibitem[{Pyeatt et~al.(2001)Pyeatt, Howe, et~al.}]{pyeatt2001decision}
L.~D. Pyeatt, A.~E. Howe, et~al., 2001. Decision tree function approximation in
  reinforcement learning. In: Proceedings of the third international symposium
  on adaptive systems: evolutionary computation and probabilistic graphical
  models. Vol.~2. Cuba, pp. 70--77.

\bibitem[{Rafiei and Ricardez-Sandoval(2018)}]{rafiei2018stochastic}
M.~Rafiei, L.~A. Ricardez-Sandoval, 2018. Stochastic back-off approach for
  integration of design and control under uncertainty. Industrial \&
  Engineering Chemistry Research 57~(12), 4351--4365.

\bibitem[{Rafiei and Ricardez-Sandoval(2020)}]{rafiei2020trust}
M.~Rafiei, L.~A. Ricardez-Sandoval, 2020. A trust-region framework for
  integration of design and control. AIChE Journal 66~(5), e16922.

\bibitem[{Rolnick et~al.(2019)Rolnick, Ahuja, Schwarz, Lillicrap, and
  Wayne}]{rolnick2019experience}
D.~Rolnick, A.~Ahuja, J.~Schwarz, T.~P. Lillicrap, G.~Wayne, 2019. Experience
  replay for continual learning.

\bibitem[{Russel et~al.(2020)Russel, Benosman, and Baar}]{russel2020robust}
R.~H. Russel, M.~Benosman, J.~V. Baar, 2020. Robust constrained-mdps:
  Soft-constrained robust policy optimization under model uncertainty.

\bibitem[{Ryu et~al.(2019)Ryu, Chow, Anderson, Tjandraatmadja, and
  Boutilier}]{ryu2019caql}
M.~Ryu, Y.~Chow, R.~Anderson, C.~Tjandraatmadja, C.~Boutilier, 2019. Caql:
  Continuous action q-learning. arXiv preprint arXiv:1909.12397.

\bibitem[{Sajedian et~al.(2019)Sajedian, Badloe, and
  Rho}]{sajedian2019optimisation}
I.~Sajedian, T.~Badloe, J.~Rho, 2019. Optimisation of colour generation from
  dielectric nanostructures using reinforcement learning. Optics express
  27~(4), 5874--5883.

\bibitem[{Satija et~al.(2020)Satija, Amortila, and
  Pineau}]{satija2020constrained}
H.~Satija, P.~Amortila, J.~Pineau, 2020. Constrained markov decision processes
  via backward value functions. arXiv preprint arXiv:2008.11811.

\bibitem[{Shin et~al.(2019)Shin, Badgwell, Liu, and
  Lee}]{shin2019reinforcement}
J.~Shin, T.~A. Badgwell, K.-H. Liu, J.~H. Lee, 2019. Reinforcement
  learning--overview of recent progress and implications for process control.
  Computers \& Chemical Engineering 127, 282--294.

\bibitem[{Singh and Kodamana(2020)}]{singh2020reinforcement}
V.~Singh, H.~Kodamana, 2020. Reinforcement learning based control of batch
  polymerisation processes. IFAC-PapersOnLine 53~(1), 667--672.

\bibitem[{Slowik and Kwasnicka(2020)}]{slowik2020evolutionary}
A.~Slowik, H.~Kwasnicka, 2020. Evolutionary algorithms and their applications
  to engineering problems. Neural Computing and Applications, 1--17.

\bibitem[{Spielberg et~al.(2019)Spielberg, Tulsyan, Lawrence, Loewen, and
  Bhushan~Gopaluni}]{spielberg2019toward}
S.~Spielberg, A.~Tulsyan, N.~P. Lawrence, P.~D. Loewen, R.~Bhushan~Gopaluni,
  2019. Toward self-driving processes: A deep reinforcement learning approach
  to control. AIChE Journal 65~(10), e16689.

\bibitem[{Sutton and Barto(2018)}]{sutton2018reinforcement}
R.~S. Sutton, A.~G. Barto, 2018. Reinforcement learning: An introduction 2nd
  ed.

\bibitem[{Sutton et~al.(2000)Sutton, McAllester, Singh, and
  Mansour}]{sutton2000policy}
R.~S. Sutton, D.~A. McAllester, S.~P. Singh, Y.~Mansour, 2000. Policy gradient
  methods for reinforcement learning with function approximation. In: Advances
  in neural information processing systems. pp. 1057--1063.

\bibitem[{Szepesv{\'a}ri(2010)}]{szepesvari2010algorithms}
C.~Szepesv{\'a}ri, 2010. Algorithms for reinforcement learning. Synthesis
  lectures on artificial intelligence and machine learning 4~(1), 1--103.

\bibitem[{Taylor et~al.(2020)Taylor, Singletary, Yue, and
  Ames}]{taylor2020learning}
A.~Taylor, A.~Singletary, Y.~Yue, A.~Ames, 2020. Learning for safety-critical
  control with control barrier functions. In: Learning for Dynamics and
  Control. PMLR, pp. 708--717.

\bibitem[{Tessler et~al.(2018)Tessler, Mankowitz, and
  Mannor}]{tessler2018reward}
C.~Tessler, D.~J. Mankowitz, S.~Mannor, 2018. Reward constrained policy
  optimization. arXiv preprint arXiv:1805.11074.

\bibitem[{Wabersich and Zeilinger(2018)}]{wabersich2018safe}
K.~P. Wabersich, M.~N. Zeilinger, 2018. Safe exploration of nonlinear dynamical
  systems: A predictive safety filter for reinforcement learning. arXiv
  preprint arXiv:1812.05506.

\bibitem[{W{\"a}chter(2002)}]{wachter2002interior}
A.~W{\"a}chter, 2002. An interior point algorithm for large-scale nonlinear
  optimization with applications in process engineering. Ph.D. thesis, PhD
  thesis, Carnegie Mellon University.

\bibitem[{Wang et~al.(2019)Wang, Li, and Chen}]{wang2019incremental}
Z.~Wang, H.-X. Li, C.~Chen, 2019. Incremental reinforcement learning in
  continuous spaces via policy relaxation and importance weighting. IEEE
  Transactions on Neural Networks and Learning Systems.

\bibitem[{Watkins and Dayan(1992)}]{watkins1992q}
C.~J. Watkins, P.~Dayan, 1992. Q-learning. Machine learning 8~(3-4), 279--292.

\bibitem[{Xie et~al.(2020)Xie, Xu, Li, Hong, and Shi}]{xie2020model}
H.~Xie, X.~Xu, Y.~Li, W.~Hong, J.~Shi, 2020. Model predictive control guided
  reinforcement learning control scheme. In: 2020 International Joint
  Conference on Neural Networks (IJCNN). IEEE, pp. 1--8.

\bibitem[{Zhou et~al.(2019)Zhou, Kearnes, Li, Zare, and
  Riley}]{zhou2019optimization}
Z.~Zhou, S.~Kearnes, L.~Li, R.~N. Zare, P.~Riley, 2019. Optimization of
  molecules via deep reinforcement learning. Scientific reports 9~(1), 1--10.

\end{thebibliography}

\end{document}